\documentclass[conference]{IEEEtran}
\IEEEoverridecommandlockouts
\usepackage{amsmath,amssymb,amsfonts}
\usepackage{booktabs}
\usepackage{cite}
\usepackage[hidelinks]{hyperref}
\usepackage{balance}
\DeclareMathOperator*{\argmin}{arg\,min}
\DeclareMathOperator*{\argmax}{arg\,max}

\title{Information Lattice Learning as Probabilistic Graphical Model Structure Learning}

\author{\IEEEauthorblockN{Haizi Yu$^{1}$ and Lav R. Varshney$^{1,2}$}
\IEEEauthorblockA{$^1$Kocree, Inc.\\
$^2$AI Innovation Institute, Stony Brook University}}

\begin{document}
\maketitle

\begin{abstract}
Information lattice learning (ILL) learns interpretable rules of a signal by alternately projecting the signal onto a partition lattice that encodes a hierarchy of abstractions and lifting selected rules back to the signal domain. When the signal is a probability mass function, we show the probabilistic rules learned by ILL admit a natural probabilistic graphical model (PGM) interpretation and develop this interpretation in detail. A partition in ILL induces a deterministic quotient variable, and a rule is the marginal law of that quotient variable. A rule set is therefore a collection of marginal constraints over interpretable abstractions. General lifting is the feasible family of all joint distributions satisfying those constraints, while special lifting chooses a maximum-ignorance reconstruction, implemented in ILL by an $L_2$ uniformity principle closely related to maximum entropy. Under a Shannon-entropy lifting, the same constraints yield a log-linear factor graph whose factors are indexed by learned abstractions. The information lattice itself, however, is not a Bayesian network: its edges encode refinement and coarsening of abstractions, not conditional dependence. Thus ILL is best viewed as structure learning for interpretable constraint-based factor graphs over quotient variables. This view clarifies how ILL relates to graphical models and maximum entropy models, while suggesting new directions for inference, identifiability, and hybrid symbolic-probabilistic learning.
\end{abstract}

%\begin{IEEEkeywords}
%Information lattice learning, probabilistic graphical models, factor graphs, maximum entropy, structure learning, interpretable machine learning.
%\end{IEEEkeywords}

\section{Introduction}
Probabilistic graphical models (PGMs) provide a language for representing distributions by exposing structure: conditional independences in Bayesian networks, Markov blankets in Markov random fields, local compatibility functions in factor graphs, and constraints in codes \cite{pearl1988,koller2009,lauritzen1996,kschischang2001,Forney2001,Forney2014}. Information lattice learning (ILL) pursues a different but adjacent objective, building on Claude Shannon's somewhat obscure lattice theory of information \cite{Shannon1953}. Rather than beginning with a graph of random variables and estimating parameters, ILL begins with a signal and searches for human-interpretable abstractions, called rules, that explain the signal with low complexity \cite{yu2023ill,yu2023illb}. The ILL approach is useful for human-interpretable knowledge discovery in scientific, artistic, and enterprise settings \cite{YuV2025}.  The ILL framework is especially useful for probabilistic modeling since the signal may itself be a probability distribution. In that setting, a rule is a mathematically precise marginal distribution over a coarsening of the sample space.

This paper develops the argument that ILL can be understood as a form of PGM structure learning, but with a crucial change in emphasis. In ordinary graphical modeling, the vertices are domain variables such as pixels, notes, atoms, or labels. In ILL, the learned objects are quotient variables induced by partitions of the signal space alphabet. These quotient variables may arise from primitive arithmetic and geometric priors, symmetry orbits, feature preimages, or lattice-theoretic joins and meets; in the group-theoretic setting, explicitly finding these quotient variables may involve solving difficult computational group theory problems \cite{YuMV2021}. A probabilistic rule says how much mass the target distribution assigns to each cell of such a partition. Thus the learned rule set is a collection of interpretable marginal constraints. The lifted distribution is the member of the corresponding constraint family that adds as little extra structure as possible.

Establishing this viewpoint has several benefits. First, it translates ILL rules into standard probabilistic language: each learned abstraction contributes a sufficient statistic or factor over the full configuration space. Second, it clarifies that although the graph, or the Hesse diagram, of an information lattice is a directed acyclic graph, it is not a Bayesian network. Its arrows express the immediate partial-order relation between abstractions, indicating that one partition is finer or coarser than another rather than conditional dependence. The lattice is therefore a structured hypothesis space from which a graphical model can be selected, not itself the probabilistic graph of the final distribution.  Finally, it enables the use of the well-established mathematical machinery of PGMs for downstream applications.

\section{ILL Rules as Marginal Laws of Quotient Variables}
Let us first start by introducing the basics of what ILL does.
Let $X$ be a finite sample space and let the signal
\begin{equation}
    \xi(x) = p(x),\qquad x\in X,
\end{equation}
be a probability mass function over $X$ rather than any generic signal. ILL defines an abstraction of $X$ as a partition $P=\{C_1,\ldots,C_m\}$ of $X$, where each cell $C_j$ is an equivalence class of configurations viewed as an abstracted concept in $P$. This representation formalizes the essence of human abstraction: abstracting away low-level details and treating distinct objects as equivalent. The projection of $p$ onto $P$ is the rule
\begin{equation}
    r_P(C)=\sum_{x\in C}p(x),\qquad C\in P. \label{eq:rule}
\end{equation}
This is the central observation for connecting ILL to PGMs, see Table \ref{tab:dictionary}. A partition $P$ induces a deterministic map
\begin{equation}
    \pi_P:X\rightarrow P,
\end{equation}
where $\pi_P(x)$ is the cell containing $x$. If $X$ is regarded as a random variable with law $p$, define the quotient variable
\begin{equation}
    Z_P=\pi_P(X).
\end{equation}
Then (\ref{eq:rule}) is precisely
\begin{equation}
    r_P(C)=\Pr\{Z_P=C\}. \label{eq:marginal}
\end{equation}
An ILL rule is therefore the marginal distribution of a deterministic abstraction of the original random object. It is neither a cluster label alone nor a parameter of a conditional distribution. It is a stochastic statement about a human-readable quotient of the configuration space.

For example, consider a piano thought experiment in ILL, where Tom and Jerry play two pianos simultaneously, each producing one note at a time. An outcome sample here is a note pair from the random variable $X=(T,J)$, where $T$ is Tom's note and $J$ is Jerry's note. A partition may distinguish whether $J>T$, giving a binary quotient variable $Z_1=\mathbf{1}\{J>T\}$. Another partition may distinguish black-key classes of $T$, giving $Z_2 = \mathbf{1}\{T \in \mathrm{BlackKeys}\}$. The corresponding rules are probabilities such as
\begin{equation}
    \Pr\{J>T\}=0.97,\qquad \Pr\{T\in \mathrm{BlackKeys}\}=0.83.
\end{equation}
They are soft symbolic regularities: probabilistic analogues of constraints such as ``Jerry usually plays higher keys'' and ``Tom usually plays black keys''.

\begin{table}[t]
\centering
\caption{Dictionary between ILL and probabilistic graphical modeling}
\label{tab:dictionary}
\begin{tabular}{@{}ll@{}}
\toprule
ILL object & PGM interpretation\\
\midrule
Signal $\xi(x)$ & Joint distribution $p(x)$\\
Partition $P$ & Deterministic abstraction $Z_P=\pi_P(X)$\\
Cell $C\in P$ & State of quotient variable $Z_P$\\
Rule $r_P(C)$ & Marginal probability $\Pr\{Z_P=C\}$\\
Rule set $R$ & Marginal-constraint family\\
General lifting $\Uparrow(R)$ & Feasible set of joint distributions\\
Special lifting $\uparrow(R)$ & Maximum-ignorance reconstruction\\
Lattice edge & Coarsening/refinement, not dependence\\
\bottomrule
\end{tabular}
\end{table}

\section{Rule Sets as Constraint-Based Graphical Models}
Let ILL select partitions $P_1,\ldots,P_k$ and associated rules $r_1,\ldots,r_k$. The learned rule set is
\begin{equation}
    R=\{r_i:i=1,\ldots,k\}.
\end{equation}
The general lifting of $R$ is the set of distributions $q$ on $X$ that satisfy every learned marginal constraint:
\begin{equation}
\begin{split}
    \mathcal{M}(R)=\{q\in \Delta_X:
    &\sum_{x\in C}q(x)=r_i(C),\\
    & C\in P_i,
    \ i=1,\ldots,k\}.
\end{split} \label{eq:feasible}
\end{equation}
Here $\Delta_X$ is the probability simplex. Equation (\ref{eq:feasible}) is the probabilistic content of multi-rule lifting. It states that the learned ILL model is not simply a list of verbal rules; it contains a convex family of joint distributions compatible with those rules.

To obtain a unique reconstruction, ILL uses a special lifting operator that chooses the most ``uniform'' member of this family. In the original formulation for computational convenience, uniformity is expressed by minimizing an $L_2$ norm subject to the rule constraints. For probability distributions, this is a convex quadratic program:
\begin{equation}
    \hat{p}_{2}=\argmin_{q\in \mathcal{M}(R)}\sum_{x\in X}q(x)^2. \label{eq:l2lift}
\end{equation}
The objective is equivalent to maximizing a Tsallis-type entropy of order two. Conceptually, (\ref{eq:l2lift}) plays the same role as a maximum-ignorance principle: respect the learned abstractions and otherwise avoid introducing gratuitous distinctions among configurations.

If one replaces (\ref{eq:l2lift}) with maximizing Shannon entropy that is also mentioned in \cite{yu2023ill}, the reconstruction becomes
\begin{equation}
    \hat{p}_{H}=\argmax_{q\in \mathcal{M}(R)} \sum_{x\in X}q(x)\log \frac{1}{q(x)}. \label{eq:maxent}
\end{equation}
The solution of (\ref{eq:maxent}), when it lies in the relative interior of the feasible polytope, has the log-linear form
\begin{equation}
    \hat{p}_{H}(x)=\frac{1}{Z(\lambda)}
    \exp\left(\sum_{i=1}^{k}\lambda_i(\pi_{P_i}(x))\right), \label{eq:loglinear}
\end{equation}
where $\lambda_i$ assigns a Lagrange multiplier to each cell of $P_i$ and $Z(\lambda)$ is the partition function from statistical physics that normalizes the distribution. Equivalently,
\begin{equation}
    \hat{p}_{H}(x)\propto \prod_{i=1}^{k}\phi_i(\pi_{P_i}(x)), \label{eq:factor}
\end{equation}
which is a factor graph with factors $\phi_i$. Each factor is constant on the cells of one learned partition. Thus, under a Shannon lifting, ILL rules generate an explicit exponential-family PGM whose sufficient statistics are indicators of interpretable partition cells. Under the $L_2$ lifting used by ILL for computational convenience, the result is not generally multiplicative in exactly this way, but it remains a convex constraint-based probabilistic model with factors or potentials indexed by learned abstractions.

\subsection{Which Graph Is Which?}
Although ILL contains a graph, i.e.,\ the Hasse diagram of an information lattice, and PGMs are also naturally graphical, they should not be conflated.

The lattice graph orders partitions by refinement. If $P\preceq P'$ means that $P$ is coarser than $P'$, then $P'$ distinguishes at least as much as $P$. The quotient variable $Z_P$ is therefore a deterministic function of $Z_{P'}$:
\begin{equation}
    Z_P=g(Z_{P'}).
\end{equation}
An edge $P\rightarrow P'$ in the lattice records an abstraction relation. It does not assert that $Z_P$ causes $Z_{P'}$, nor that a conditional independence relation holds. This is why ILL should not be read directly as a Bayesian network over rules.

The PGM graph is instead induced after a rule set is chosen. In a factor-graph view, the variable side contains the ground degrees of freedom needed to specify $x\in X$, while the factor side contains selected abstractions $P_i$ or quotient variables $Z_{P_i}$. A factor node $\phi_i$ connects to precisely those ground variables on which $\pi_{P_i}$ depends. When $\pi_{P_i}$ is a simple coordinate feature, this is a local factor. When $\pi_{P_i}$ is a symmetry orbit, ordering relation, modular arithmetic feature, or join-generated abstraction, the factor may be global but remains interpretable.

This distinction yields a useful two-graph semantics:
\begin{enumerate}
    \item the \emph{lattice graph} is a search graph over candidate abstractions;
    \item the \emph{probabilistic graph} is a factorization or constraint graph induced by the selected abstractions.
\end{enumerate}
The first graph supports explanation and rule discovery, whereas the second supports probabilistic reconstruction and inference.

\subsection{Example: Two Abstract Rules}
Consider a finite ground space $X=\{1,\ldots,6\}$, interpreted as die outcomes, and let $p$ be an unknown distribution. Suppose ILL selects two partitions
\begin{equation}
P_1=\{\{1,3,5\},\{2,4,6\}\},\qquad
P_2=\{\{1,2,3,4\},\{5,6\}\}.
\end{equation}
The first quotient variable is parity, and the second distinguishes small from large outcomes. A rule set such as
\begin{equation}
\Pr\{Z_{P_1}=\mathrm{odd}\}=0.6,\qquad
\Pr\{Z_{P_2}=\mathrm{small}\}=0.7
\end{equation}
imposes two marginal constraints:
\begin{align}
q_1+q_3+q_5&=0.6,\\
q_1+q_2+q_3+q_4&=0.7.
\end{align}
Many full distributions satisfy these equations. Lifting chooses one of them according to a uniformity principle. In the Shannon version, there are two learned factors,
\begin{equation}
\hat p(x)\propto \phi_1(\mathrm{parity}(x))\phi_2(\mathrm{size}(x)).
\end{equation}
The graph has one ground variable and two factor nodes, but the meaningful content is not the graph alone; it is the fact that the factor labels, parity and size, are interpretable abstractions selected from the lattice.

This example also shows the role of the join. The join $P_1\vee P_2$ partitions outcomes by the joint abstraction $(Z_{P_1},Z_{P_2})$, e.g., odd-small, even-small, odd-large, and even-large. The two rules specify only the two one-dimensional marginals of this joint abstraction. Lifting fills in the missing association between parity and size. If the original signal exhibits no interaction beyond the learned constraints, the lifted model recovers it well. If the original signal contains a strong parity-by-size interaction, ILL must select an additional rule on the join or on another abstraction that captures the residual structure.

\section{Antichains, Redundancy, and Structure Learning}
ILL searches for simple rule sets that reconstruct the signal well \cite{YuV2017}. In the practical learning phase, the search is restricted to antichains in the information sublattice. This has a direct probabilistic meaning. If one partition refines another, then the finer quotient variable determines the coarser one. Including both rules can be redundant: the marginal of a coarse quotient can often be computed from the marginal or joint information available at a finer quotient. Antichain selection discourages such deterministic redundancy and favors rule sets that capture distinct explanatory directions.

From a PGM perspective, ILL is therefore a form of structure learning. But it is not ordinary edge selection among a fixed set of variables \cite{Choi,Liu}. The candidate structures are quotient variables constructed by prior-driven mechanisms: feature preimages, symmetry orbits, and lattice joins or meets. The learning algorithm evaluates them by their explanatory contribution under projection and lifting. A rule is valuable when adding its marginal constraint substantially improves the lifted reconstruction of the original distribution. A rule is simple when its entropy is small and when it can be described by a simple partition tag.

This differs from conventional maximum-likelihood structure learning in three ways. First, the objective is explanatory reconstruction rather than predictive likelihood alone. Second, the candidate factors are not arbitrary basis functions but interpretable abstractions. Third, lattice order supplies a non-statistical notion of redundancy: refinement and coarsening relations identify rules that are logically or deterministically related even before data are observed.

\subsection{Model Selection Criteria}
The PGM view further clarifies what should be optimized during learning. Let $R$ be a candidate antichain. Its goodness can be decomposed into at least four terms:
\begin{equation}
\mathcal{L}(R)=D(p\,\|\,\uparrow R)+\alpha |R|+\beta\sum_{r\in R}H(r)+\gamma\Omega(R),
\end{equation}
where $D$ measures reconstruction loss, $|R|$ penalizes the number of learned abstractions, $H(r)$ penalizes high-entropy rules, and $\Omega(R)$ penalizes semantic or computational complexity of the partition descriptions. Here $\alpha, \beta$, and $\gamma$ are tradeoff parameters.  The first term is familiar from probabilistic modeling; the remaining terms encode interpretability. This makes ILL different from pure likelihood-based graphical-model learning, where a high-order factor may be accepted if it improves likelihood even when it is unintelligible. In ILL, a factor must also be a good explanation.

There is also a natural Bayesian variant. One may place a prior over partitions favoring simple tags, low description length, low entropy, and antichain diversity, then infer a posterior over rule sets. Such a Bayesian ILL would produce not one explanation but a distribution over explanations. This is appealing when several antichains reconstruct the signal almost equally well. The posterior mass would quantify ambiguity in the explanatory structure rather than merely uncertainty in numerical parameters.

\subsection{Relation to Existing PGM Families}
The ILL-as-PGM view is closest to maximum entropy modeling and factor graphs. Maximum entropy models define a distribution by matching specified expectations while maximizing entropy \cite{jaynes1957}. ILL similarly defines a distribution by matching selected rule marginals and choosing a uniform reconstruction. Factor graphs provide a natural representation for the resulting product of learned abstraction factors in the Shannon lifting case \cite{kschischang2001,wainwright2008}.

The relation to Bayesian networks is weaker. A Bayesian network factorizes a joint distribution into conditional probability tables according to a directed acyclic graph. ILL's lattice is also a directed acyclic graph, but the edges are abstraction-order relations. Unless an additional causal or conditional-dependence semantics is imposed, the lattice does not entail the independences that would make it a Bayesian network. A selected ILL rule set may be converted to a Bayesian network only after choosing an ordering and introducing quotient variables as deterministic children or parents of ground variables; that conversion is auxiliary rather than intrinsic.

The relation to Markov random fields is stronger. If each learned abstraction depends on a subset of ground variables, then the Shannon-lifted model in (\ref{eq:factor}) is an undirected factorization over those subsets. If an abstraction is global, the corresponding factor is high-order. Thus ILL can be understood as learning high-order, human-readable potentials. This is important because many scientific rules are naturally high-order: chordal relations in music, valence constraints in chemistry, conservation laws in physics, and spatial symmetries in images are not always expressible as low-order pairwise interactions.

\section{Inference and Learning in the PGM View}
Projection and lifting become the two fundamental inference operations. Projection maps a joint model to an abstract marginal:
\begin{equation}
    q(x)\mapsto q_{P}(C)=\sum_{x\in C}q(x).
\end{equation}
Lifting solves the inverse problem: given several abstract marginals, reconstruct a compatible joint distribution. In ordinary graphical modeling, inference usually means computing marginals from a specified joint model. In ILL, inference alternates between computing abstract marginals and reconstructing the least-committal joint model consistent with selected marginals.

This inversion changes the role of factors. In a factor graph, factors are often treated as primitive local compatibilities from which a joint law is assembled. In ILL, factors are learned as summaries of a target signal. The lifted distribution is then the joint law implied by those summaries under a uniformity principle. The learned factors are thus closer to empirical sufficient statistics or moment constraints.

The approach also clarifies what it means for ILL rules to be probabilistic rather than symbolic. A hard symbolic rule corresponds to a quotient variable with probability one on a cell, such as $\Pr\{J>T\}=1$. A soft ILL rule permits exceptions, such as $\Pr\{J>T\}=0.97$. Low entropy rules are nearly deterministic and hence easy to interpret; higher entropy rules may still be useful but are less rule-like. This explains why entropy appears as a simplicity measure in ILL and as a measure of determinism for learned abstractions.

\subsection{Semantics of Explanation}
The PGM interpretation also explains why ILL rules are useful for human understanding. A conventional graphical model may represent a distribution compactly while leaving its variables semantically opaque. ILL reverses this priority. It constructs candidate variables from mechanisms intended to be cognitively and scientifically meaningful, then asks which of those variables carry enough probability mass structure to explain the signal. The resulting model is not just compressed; it is compressed along named equivalence relations that encode interpretable concepts.

This suggests a three-level semantics. At the first level, a rule is a statistical object: a marginal distribution of $Z_P$. At the second level, it is an algebraic object: a constraint on the probability simplex associated with a partition. At the third level, it is a semantic object: a statement about the domain criterion that generated the partition. For instance, a rule may be simultaneously a binary marginal, a half-space constraint over probabilities, and a musical statement about relative pitch. The explanatory power of ILL comes from maintaining all three levels at once, cf.~\cite{Weaver1949}.

\section{Conclusion and Discussions}

The probabilistic rules learned by ILL are understood as marginal laws of deterministic quotient variables. A learned rule set is a family of marginal constraints, and lifting is a maximum-ignorance reconstruction of a joint distribution satisfying those constraints. Under the Shannon entropy lifting, this reconstruction is a log-linear factor graph with factors indexed by learned abstractions; under the $L_2$ lifting often used in ILL for computational convenience, it remains a convex, constraint-based probabilistic model with a closely related uniformity principle. The information lattice is not itself a Bayesian network, because its edges encode the coarsening--refinement relation rather than conditional dependence. Instead, the lattice is a structured hypothesis space for selecting interpretable factors. In this sense, ILL provides a distinctive bridge between symbolic rule learning and probabilistic graphical modeling: it learns not only parameters of a distribution, but the human-readable abstractions through which the distribution should be explained.

\subsection{Scope of the PGM Analogy}
The relationship between ILL and PGMs is powerful but there are some caveats. ILL's Tsallis entropy lifting does not generally yield the same multiplicative factorization as Shannon maximum entropy, except in the special case of the entropy index $q$ approaching $1$ when the Tsallis entropy coincides with the Shannon entropy. When the entropy index is $q = 2$, it is better interpreted as a quadratic projection onto a linear constraint set, while the Shannon version is an exponential-family projection. One should also note that ILL partitions need not be local in the coordinate sense (and in fact inducing long-distance invariances is the great power of the ILL approach). A symmetry-induced abstraction may depend on all coordinates of $x$, so the induced factor graph may contain high-order factors that are expensive for standard message passing. Finally, one should note that the learned rules are observational summaries that have mechanistic causability. Without additional assumptions, they do not specify causal mechanisms or intervention responses as may be needed in causal inference.

These limitations mark the boundary between two modeling traditions. PGMs traditionally emphasize tractable factorization and inference. ILL emphasizes interpretable abstraction and rule discovery. As such, it is fruitful to use graphical-model theory to analyze the probabilistic consequences of selected ILL rules.

\subsection{Future Work and Applications}
Viewing ILL rules as PGM factors suggests several research directions.

First is statistical estimation: Given finite samples, the empirical projected rule $\hat r_P$ has sampling error. Confidence regions for rule marginals would induce uncertainty sets in the lifted distribution, yielding robust ILL models.

Second is identifiability: Different antichains can define the same lifted distribution, especially when partitions have nontrivial joins and deterministic relations. The PGM view turns this into an equivalence problem among constraint sets and exponential-family sufficient statistics. Characterizing minimal identifiable antichains would strengthen the theory of ILL explanations.
Notably, compared to recent world-model identifiability work that recovers underlying latent variables up to a transformation \cite{KlindtLB2026}, ILL identifies the world model as the closure of the discovered rules and abstractions, thereby recovering the underlying compositional system of rules of the world rather than a particular latent coordinate system.

Third is inference: General lifting can be expensive when the ground space is large. Factor-graph algorithms, variational approximations, and message passing may become available once selected ILL rules are expressed as factors. Conversely, lattice structure may suggest new message-passing schemes over abstraction hierarchies rather than over coordinate variables alone.

Fourth is causality: ILL rules can resemble causal laws and in fact have large causability due to their mechanistic explanations \cite{yu2023ill}, but the lattice alone does not provide intervention semantics. To obtain causal PGMs, one would need to distinguish observational abstractions from mechanisms stable under intervention. This suggests an approach where ILL proposes interpretable quotient variables and causal modeling tests whether they support invariant mechanisms.

Finally is formal verifiability: The PGM representation of ILL rules can be used to construct finite state machine representations-based digital twins of systems that can be represented in Lamport's temporal logic of action \cite{Lamport}, such that they can be formally verified \cite{Huang}.  

Given that ILL rules achieve optimal lossy semantic compression by thinking of it as a form of abstraction (together with group source codes) \cite{YuV2025a}, and performant forms of co-creativity that take abstractions of knowledge and recombine them into new compositions \cite{YuEGKPV2021}, it is of interest to interpret compression and creativity in terms of changes to PGMs. Finally, ILL methods achieve state-of-the-art performance for visual classification \cite{YuMVE2025}, and again it is of interest to interpret the underlying semantic rules that emerge in terms of graphical models \cite{Willsky}.

\end{document}